%% file: main.tex
\documentclass[10pt,twocolumn,letterpaper]{article}

\usepackage{iccv}
\usepackage{times}
\usepackage{epsfig}
\usepackage{graphicx}
\usepackage{amsmath}
\usepackage{amssymb}
\usepackage{multirow}
\usepackage{mathtools}
\usepackage{tabularx}
\usepackage{xcolor}
\usepackage{booktabs}
\usepackage[accsupp]{axessibility} 

\usepackage[pagebackref=true,breaklinks=true,colorlinks,bookmarks=false]{hyperref}

\iccvfinalcopy 

\def\iccvPaperID{1895} 

\ificcvfinal\fi

\begin{document}
\newcommand{\model}{UnLoc}
\title{\model{}: A Unified Framework for Video Localization Tasks}

\author{Shen Yan$^1$\thanks{Equal contribution.} \;\; Xuehan Xiong$^1$\footnotemark[1] \;\; Arsha Nagrani$^1$ \;\; Anurag Arnab$^1$ \;\; Zhonghao Wang$^2$\thanks{Work done while an intern at Google}\;\; \\
Weina Ge$^1$\;\; David Ross$^1$ \;\; Cordelia Schmid$^1$ \\
$^1$Google Research\quad
$^2$University of Illinois at Urbana-Champaign \\
{\tt \small \{shenyan, xxman, anagrani, aarnab, dross, cordelias\}@google.com \;\; \{wzhonghao95\}@gmail.com \;\;}\\
}

\maketitle
\ificcvfinal\thispagestyle{empty}\fi

\begin{abstract}
\input{abstract.tex}
\end{abstract}

\input{introduction.tex}

\input{related_work.tex}

\input{method.tex}
\input{experiments.tex}
\input{conclusion.tex}
\input{ack.tex}

{\small
\bibliographystyle{ieee_fullname}
\bibliography{bibliography}
}
\end{document}

%% file: abstract.tex
While large-scale image-text pretrained models such as CLIP have been used for multiple video-level tasks on trimmed videos, their use for temporal localization in untrimmed videos is still a relatively unexplored task. We design a new approach for this called \model{}, which uses pretrained image and text towers, and feeds tokens to a video-text fusion model. The output of the fusion module are then used to construct a feature pyramid in which each level connects to a head
to predict a per-frame relevancy score and start/end time displacements. Unlike previous works, our architecture enables Moment Retrieval, Temporal Localization, and Action Segmentation with a single stage model, without the need for action proposals, motion based pretrained features or representation masking. Unlike specialized models, we achieve state of the art results on all three different localization tasks with a unified approach. Code will be available at: \url{https://github.com/google-research/scenic}.

%% file: introduction.tex
\section{Introduction}






Contrastive vision-language pretraining has been shown to learn powerful feature representations, and moreover
enables open-set inference on a wide range of tasks~\cite{radford2021learning,jia2021scaling}.
As a result, pretrained models such as CLIP~\cite{radford2021learning} have been adapted to multiple diverse tasks including video classification~\cite{ni2022expanding,lin2022frozen}, object detection~\cite{minderer2022simple} and segmentation~\cite{ghiasi2022scaling}.

In this paper, we study how to adapt large-scale, contrastively trained image-text models to untrimmed video understanding tasks that involve localization. While CLIP has been used widely for trimmed video tasks (classification~\cite{ni2022expanding,lin2022frozen} or retrieval~\cite{bain2022clip}), its use on long, untrimmed video is still in a nascent stage. Long videos come with multiple challenges -- CLIP is pretrained on images only, and localization in untrimmed videos requires exploiting \textit{finegrained} temporal structured information in videos. In particular, it is challenging for image and language models to learn properties
of temporal backgrounds  (with respective to foreground actions) during training. In contrast, natural videos often come with a large, variable proportion of background and detecting specific actions is critical for localization tasks~\cite{nag2022zero}. Finally, localization in long untrimmed videos also typically involves detecting events at multiple temporal scales. Consequently, existing approaches that use CLIP typically focus on a two-stage approach involving off-the-shelf proposal generators~\cite{ju2022prompting}, or use temporal features such as I3D~\cite{mun2020local} or C3D~\cite{soldan2021vlg}.
In contrast, we propose an end-to-end trainable one-stage approach starting from a CLIP two tower model only.

\input{teaser.tex}
We focus specifically on three different video localization tasks - Moment Retrieval (MR)~\cite{krishna2017dense,gao2017tall}, Temporal Action Localization (TAL)~\cite{heilbron2015activitynet,idrees2017thumos} and Action Segmentation (AS)~\cite{tang2019coin}.
These tasks have typically been studied separately, with different techniques proposed for each task.
We show how we can use a single, unified approach, to address all of these tasks, without using any external proposals. We do this by leveraging a two-tower model (with a vision and text encoder), in conjunction with a single video-text fusion module, which performs mid-level fusion of text and visual tokens (Figure~\ref{fig:teaser}). Our two tower model can naturally handle tasks such as moment retrieval which contain both video and text as input modalities, and can be used for open-set inference in other tasks such as temporal action localization and action segmentation. While many works use the visual encoder only~\cite{buch2017bmvc,chao2018rethinking,girdhar_cvpr_2019,zhang2022actionformer}, we believe that the language priors learnt with the pretrained text encoder can contain useful information and should be leveraged together with the image encoder early in the model design (particularly for open-set evaluation), and not right at the end for similarity computation. Inspired by existing object detection works~\cite{li2022exploring}, we also use the output frame tokens from our fusion module to construct a feature pyramid, to enable understanding at multiple temporal scales.  


Our approach achieves state-of-the-art results across all three video localization tasks - MR~\cite{krishna2017dense,gao2017tall}, TAL~\cite{heilbron2015activitynet,idrees2017thumos} and AS~\cite{tang2019coin}.
We also perform thorough ablation studies, studying the effect of modelling choices across a range of tasks.

%% file: teaser.tex
\begin{figure}[t]
\centering
\includegraphics[width=1.0\linewidth]{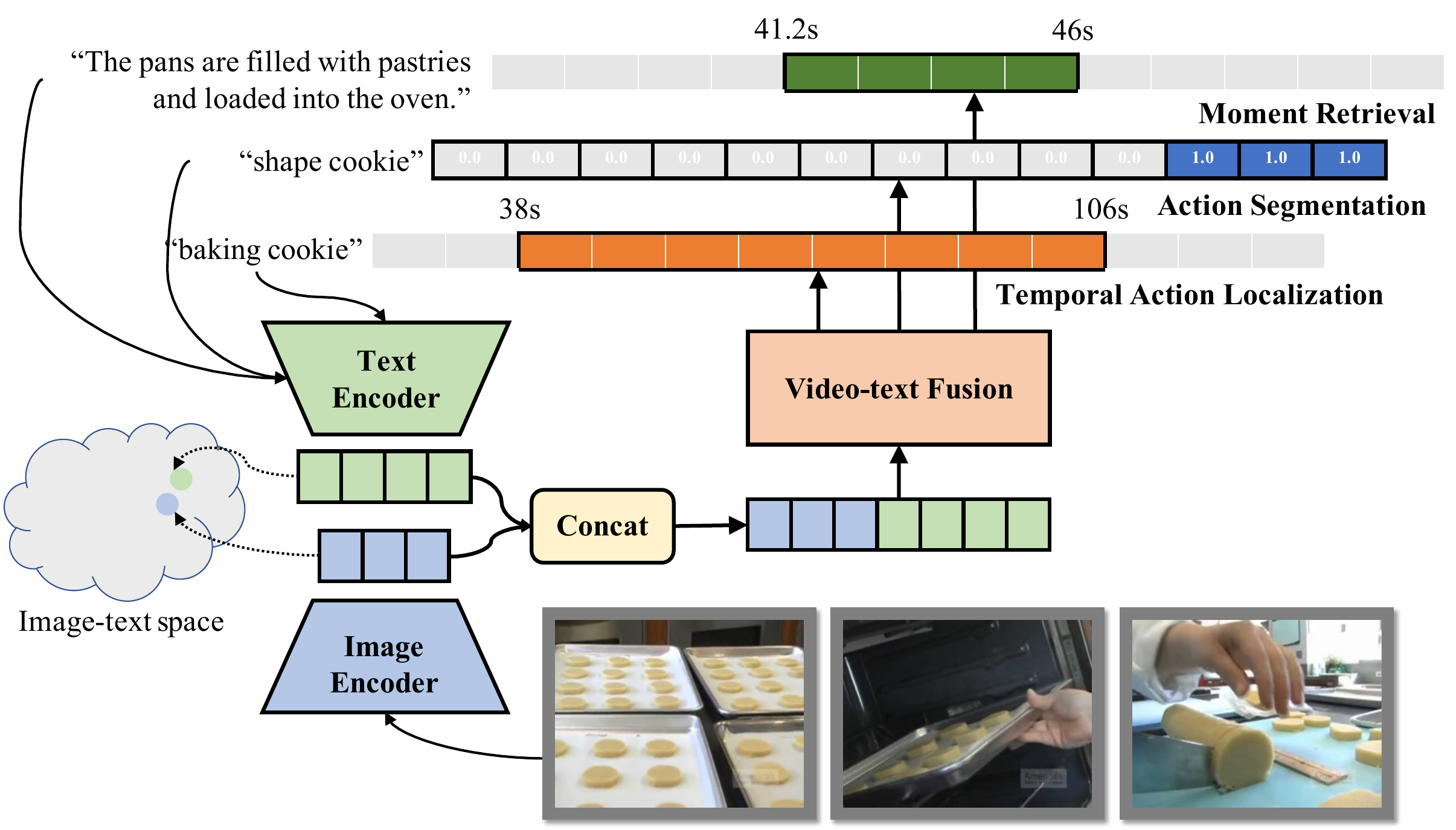}
    \caption{\textbf{Applying two-tower CLIP to video localization tasks:} We propose \textbf{\model}, a single stage, unified model that achieves state of the art results on 3 different video localization tasks - moment retrieval, temporal action localization and action segmentation. \model{} leverages a two-tower model (with a vision and text encoder) in conjunction with a video-text fusion module and feature pyramid to perform mid-level feature fusion without the need for any temporal proposals.}
    \label{fig:teaser}
\end{figure}

%% file: related_work.tex
\section{Related Work}

\textbf{Models based on CLIP for localization.} 
Most works use CLIP for video level tasks that operate on short, trimmed clips, for example for classification tasks (\emph{e.g.} ActionCLIP~\cite{wang2021actionclip} and X-CLIP~\cite{ni2022expanding}). EVL~\cite{lin2022frozen} also adapts CLIP to video classification, but does so by training a small number of extra parameters. CLIP has also been used for other video level tasks such as text-video retrieval, as done in CLIP4CLIP~\cite{Luo2021CLIP4Clip} and CLIP-Hitchikers~\cite{bain2022clip}. A number of works also use CLIP for tasks such as object detection~\cite{minderer2022simple,zhong2022regionclip} and segmentation~\cite{ghiasi2021scaling, ding2021decoupling}. Works that use CLIP for localization tasks in untrimmed videos, on the other hand are less common. Vid2Seq~\cite{yang2023vid2seq} uses CLIP features for dense video captioning, where temporal boundaries and captions are predicted together. Most works for localization however still reply heavily on I3D~\cite{carreira_cvpr_2017}, C3D~\cite{tran_iccv_2015}, R(2+1)D~\cite{tran_cvpr_2018}, VGG~\cite{simonyan2014very}, or SlowFast~\cite{feichtenhofer_iccv_2019} features for moment retrieval~\cite{soldan2021vlg, zeng2020dense, zhang2020learning, zhang2020span,nan2021interventional}, temporal action localization~\cite{nag2022zero,zhu2021enriching,zhang2022actionformer} and action segmentation~\cite{luo2020univl}.   

\textbf{Temporal Action Localization (TAL).} 
Supervised learning-based temporal action localization can be summarized into two-stage~\cite{shou2016temporal,chao2018rethinking,SSN2017ICCV,lin2019bmn,lin2018bsn} and single-stage methods~\cite{buch2017bmvc,lin2017sst,nawhal2021activity,zhang2022actionformer}. More recently, EffPrompt~\cite{ju2022prompting} uses a two-stage sequential localization and classification architecture for zero-shot action localization, with the first stage consisting of action proposal generation with an off-the-shelf pre-trained proposal detector (e.g., BMN~\cite{lin2019bmn}), followed by proposal classification using CLIP features. We aim to build a proposal-free framework and directly regress the temporal location of the corresponding class labels or queries by using the fused video-text features. The closest to our method is STALE~\cite{nag2022zero}, which trains a single-stage model for zero-shot localization and classification, using representation masking for frame level localization. Unlike STALE, which evaluates on only TAL, we present a single unified method for MR, TAL and AS, and also introduce a feature pyramid for multi-scale reasoning.

\textbf{Moment Retrieval (MR).} 
Unlike TAL, where class names are predefined used a closed-form vocabulary, MR aims to find the relevant clip in 
an untrimmed video for a given open-ended natural language query.
Early works use sliding windows over video sequences to generate video segment proposals~\cite{hendricks2017localizing,gao2017tall}, after which the proposals are ranked by their similarity to the query. This ignores the fine-grained relationships between video frames and the words in sentences. Anchor-based methods~\cite{chen-etal-2018-temporally,wang2019temporally,yuan2019semantic} avoid proposal generation by assigning each frame with multi-scale anchors sequentially and use these to obtain more fine-grained matchings between video and text. Regression-based methods~\cite{chen2020rethinking,zeng2020dense,zhang2020learning,mun2020local,lei2021detecting,liu2022umt} involve learning cross-modal interactions to directly predict the temporal boundaries of a target moment without the need for proposal generation. Our work belongs to this category, unlike works that tend to use the text tower only at the end to compute similarity scores \cite{hendricks2018temp,zhang2020hierarchical,gao2021,zhang2020learning,liu2022umt}, we fuse image and text tokens early on in our model to better leverage language priors from the pretrained CLIP text tower.  

\textbf{Action Segmentation (AS).}
Action segmentation involves assigning a pre-defined label to each token or frame in a untrimmed long video, which helps to distinguish meaningful video segments from other tokens or frames~\cite{tang2019coin}. While previous works~\cite{sun2019learning,miech2020end,xu-etal-2021-videoclip,luo2020univl} pretrained their models on HowTo100M~\cite{miech2019howto100m}, our approach involves initializing models with pretrained CLIP models. CLIP was trained on pairs of web images and text, which may be less prone to noise compared to ASR and clip pairs.

%% file: method.tex
\section{Method}

%
Our model unifies three tasks: MR, TAL and AS, which we first define in Sec. \ref{sec:tasks}.
As shown in Fig.~\ref{fig:overview}, our model (Sec.~\ref{sec:unified_architecture}) first tokenizes a (video, text) pair and then fuses information from the two modalities together with a simple video-text fusion module.
To capture the multi-scale information needed for localization, we then construct a Feature Pyramid (Sec.~\ref{sec:fpn}) on the output of the video-text fusion module.
These multi-scale features are then fed into a task-specifc Head module (Sec.~\ref{sec:head}) to localize activities or ``ground'' a language description.

\input{method_overview.tex}
\subsection{Tasks}
\label{sec:tasks}
\emph{Moment Retrieval (MR)}, also known as Video Grounding, is the task of matching a given language description (query) to specific video segments in an untrimmed video. 
\emph{Temporal Action Localization (TAL)}  aims to detect events in a video and output the corresponding start- and end-timestamps. One key difference from MR is that events in TAL are from a predefined closed-vocabulary set, often described by a short phrase (e.g., ``baking cookies''). 
Finally, similar to Semantic Segmentation, which parses images into semantic categories at a pixel level, \emph{Action Segmentation (AS)} involves producing activity labels at a frame level. Also, for this task the labels are typically predefined from a closed-vocabulary set.

\subsection{A unified architecture}
\label{sec:unified_architecture}
Our model takes (video, text) pairs as inputs, and for each frame in the video it outputs a relevancy score between the frame and the input text, as well as the time differences between the frame and the start/end timestamps of the predicted segment. The target relevancy score is set to 1 if a frame falls within the labeled segment, otherwise zero. In the case of TAL and AS, we use class labels as the input texts while in MR, text queries are used as input texts. For each video we form $C$ (video, text) pairs where $C$ is the number of classes in TAL and AS and for MR $C$ is the number of captions associated with this video.

Fig.~\ref{fig:overview} gives an overview of our proposed architecture. The input pair is first tokenized and encoded by a pair of image and text encoders. The two encoders are initialized from a pretrained, CLIP visual language model~\cite{radford2021learning}. The two encoders come from the same pretrained model -- pretrained by aligning image and text pairs with a contrastive loss, which provides a strong prior on measuring the relevancy between each frame and the input text. This is one of key contributing factors to the success of our model. As Sec.~\ref{sec:ablations} will show, using ``unpaired'' image/text encoders indeed diminishes the performance. 

After tokenization and encoding, the input video and text are represented by $N$ frame tokens and $T$ text tokens. We then form a new sequence by concatenating $N$ frame tokens with either a single token (e.g., CLS) representing the whole text sequence or all $T$ text tokens (See Sec.~\ref{sec:ablations} for ablation). The concatenated sequence is then fed into a video-text fusion module. In this work, we implement this fusion module using a transformer encoder~\cite{vaswani_neurips_2017,devlin_naacl_2019}. This encoder performs two key functions -- (i) it is a temporal encoder, able to model inter-frame correspondences omitted by the image-only CLIP model, and (ii) it can also function as a refinement network, with the ability to correct mistakes made by the CLIP model.  After fusion, only frame tokens $\mathbf{X}^{c} \in \mathbb{R}^{N \times K}$ are used to construct a feature pyramid where each level is created by downsampling the original sequence using strided convolutions where $c$ is the index of the class or caption and $K$ denotes the hidden size of the token. This process is repeated for all class labels/captions. Text tokens are omitted from this construction because their information has been incorporated into the frame tokens by the fusion module, and they do not correspond to any timestamps. 

Finally, each pyramid level connects to a Head module to predict a per-frame relevancy score, $\mathbf{\hat{y}}_l^c \in \mathbb{R}^{N_l}$, and start/end time displacements, $\mathbf{\hat{t}}_l^c \in \mathbb{R}^{N_l \times 2}$, where $N_l$ denotes the number of features in pyramid level $l$. The final number of predictions is $\sum_{i=1}^{L} \frac{N}{2^{i-1}}$, and is therefore greater than $N$ if there is more than one level in the feature pyramid.
For example, if we construct a 3-level feature pyramid the total number of predictions will be $N + N/2 + N/4$. Each prediction is expanded into a temporal segment by applying the predicted displacements to its frame timestamp. Given these temporal segments for all pyramid layers,
we filter out overlapping segments during inference with soft non-maximal suppression (SoftNMS)~\cite{bodla2017soft}.

\subsection{Feature pyramid}
\label{sec:fpn}
A feature pyramid can improve a model's capability to detect events at different scales. For example, features from the top level can detect events with a long duration while bottom-level features can localise short segments. 
Feature Pyramid Networks (FPN~\cite{lin2017feature}) have been used extensively in object detection for images to pass richer semantic information from a higher level in the CNNs to lower level feature maps that have higher spatial resolution. We propose another simpler structure inspired by ViTDet~\cite{li2022exploring} by removing the lateral and top-down connections in the FPN. Since the last layer in the transformer encoder contains the most semantic information~\cite{raghu2021vision} 
 and shares the same temporal resolution as the first one, the lateral and top-down connections are no longer required. The feature pyramid is constructed by applying convolution with different strides to the output tokens from the last transformer layer in the video-text fusion module (See Fig.~\ref{fig:overview}a). Note that text tokens are not used during the feature pyramid construction since their information has been fused into the frame tokens. This simpler design removes the downsampling step in the encoder and allows us to share the same architecture used in pretraining stage (See Sec.~\ref{sec:implementation} for more details). 
Similar to findings in ~\cite{li2022exploring}, our ablation (Sec.~\ref{sec:ablations}) shows that this simpler design outperforms FPN on TAL as it introduces less additional layers to the pretrained model. AS is a frame-level task so features from only the bottom level in the feature pyramid are used for prediction.

\subsection{Head design}
\label{sec:head}
As shown in Fig.~\ref{fig:overview}b, we have two heads, one for relevancy score prediction and the other for displacement regression. Although the two heads share the same structure their weights are not shared. Our head design following~\cite{zhang2022actionformer} is simple consisting of $M$ 1D convolution blocks where each block is made of three operations: Layer Normalization~\cite{ba_arxiv_2016}, 1D convolution, and a ReLU activation~\cite{fukushima1975cognitron}. A convolution (e.g., a local operation) is used to encourage nearby frames to share the same label.
At the end of each head, a linear layer is learned to predict per-frame relevancy scores $\mathbf{\hat{y}}^{c} \in \mathbb{R}^{N\times 1}$ or to predict per-frame start/end time displacements $\Delta\mathbf{\hat{t}}^{c} \in \mathbb{R}^{N \times 2}$:
%
\begin{align}
\mathbf{\hat{y}}^{c} &= \mathbf{Z}^{c}\mathbf{w}_{cls} + b_{cls} \label{eq:cls_head} \\
\Delta\mathbf{\hat{t}}^{c} &= \text{relu}(\mathbf{Z}^{c}\mathbf{w}_{reg} + \mathbf{b}_{reg}) \label{eq:reg_head}
\end{align}
where $\mathbf{Z}^{c}$ are the activations of frame tokens $\mathbf{X}^{c}$ after convolution blocks, $\mathbf{w}_{cls} \in \mathbb{R}^{K\times 1}$ and $b_{cls} \in \mathbb{R}^{1\times 1}$ are the weights and bias for the classification head, and $\mathbf{w}_{reg} \in \mathbb{R}^{K\times 2}$ and $\mathbf{b}_{reg} \in \mathbb{R}^{1\times 2}$ are the weights and biases for the regression head.
We limit the predicted displacements to be greater or equal to zero through a ReLU non-linearity. Eqs.~\ref{eq:cls_head} and~\ref{eq:reg_head} are repeated to generate scores and displacements for every class/caption and the same learned weight and bias terms are shared.
For AS only the relevancy scoring head is used. One key difference from~\cite{zhang2022actionformer} is that our model predicts a different start/end time displacement for each class while ~\cite{zhang2022actionformer} predicts one displacement $\Delta\mathbf{\hat{t}} \in \mathbb{R}^{N\times 2}$ shared among all classes, which assumes that there is no overlapping segment in the video.

\subsection{Loss function}
\label{sec:loss}
For AS, we use sigmoid cross entropy loss to measure the relevance between a frame and class label. For TAL and MR, we use the focal loss~\cite{lin2017focal} for the relevancy scoring head as class imbalance is a known issue in one-stage detectors~\cite{lin2017focal}. 
For the regression head we experiment with four popular regression losses, L1, IoU, DIoU~\cite{zheng2020distance}, and L1+IoU. The L1 loss computes the absolute distance between the predicted and the ground truth start/end times.
The IoU loss directly optimizes the intersection of union objective, which is defined as
\begin{align*}
L_{iou} &= 1 - IoU(\Delta\hat{s}, \Delta\hat{e}) \\
        &= 1 - \frac{\min(\Delta\hat{s}, \Delta s)+\min(\Delta\hat{e}, \Delta e)}{\max(\Delta\hat{s}, \Delta s)+\max(\Delta\hat{e}, \Delta e)}
\end{align*}
where $\Delta\hat{s}, \Delta\hat{e}$ and $\Delta s, \Delta e$ are the predicted and the ground truth displacements to the start/end times. If $\Delta\hat{s}$ or $\Delta\hat{e}$ is zero, its gradient will also be zero, which could happen due to poor initialization. Distance IoU (DIoU~\cite{zheng2020distance}) is proposed to address the zero-gradient issue by also taking into account the distance between the two centers of the ground truth box and the predicted box.
We end up using L1 loss based on the ablation in Sec.~\ref{sec:ablations} and also apply a weight factor $\alpha$ to balance between the focal loss and L1 loss.

%% file: method_overview.tex
\begin{figure*}[t]
    \centering
    \includegraphics[width=0.9\linewidth]{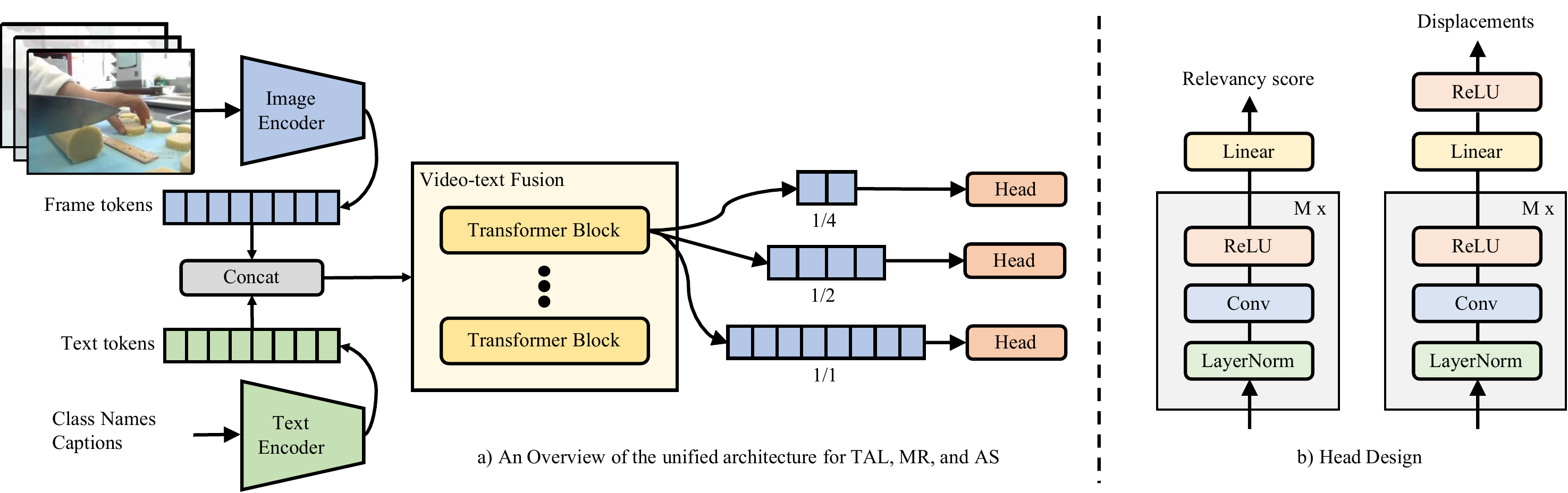}
    \caption{\textbf{Overview of our method \model{}.} Given a video and text (e.g., class names in TAL/AS or captions in MR) pair, first they are tokenized and encoded by a pair of image and text encoders. Frame and text tokens are concatenated into a long sequence and then fed into a transformer for fusion. Frame tokens from the last transformer layer are used to construct a feature pyramid in which each level connects to a head to predict a per-frame relevancy score and start/end time displacements. No text token is used to construct the feature pyramid since text information has already been ``fused'' into the frame tokens via self-attention. We show a 3 layer feature pyramid for simplicity. All heads across different pyramid levels share the same weights.}
    \label{fig:overview}
\end{figure*}

%% file: experiments.tex
\section{Experimental Evaluation}
We first describe datasets, evaluation metrics and implementation details in Sec.~\ref{sec:exp_setup}. We then provide a number of ablations on our architecture design, use of the text encoder, video-text fusion module and finetuning strategies (Sec.~\ref{sec:ablations}). Finally, we show the results of our method compared to the state-of-the-art in Sec~\ref{sec:sota}.
\subsection{Experimental setup} \label{sec:exp_setup}
\subsubsection{Datasets and evaluation metrics}
\textbf{Moment retrieval.} \emph{ActivityNet Captions}~\cite{krishna2017dense} contains 20,000 videos and 100,000 segments where each is annotated with a caption by human. On average each caption contains 13.5 words and videos have an average duration of ~2 minutes. The dataset is divided into three splits, train, val\_1, and val\_2. Following ~\cite{zeng2020dense,soldan2021vlg} we use train split for training, val\_1 for validation and val\_2 for testing. \emph{Charades-STA}~\cite{gao2017tall} contains 6,672 videos and 16,128 segment/caption pairs, where 12,408 pairs are used for training and 3720 for testing. Each video is annotated with 2.4 segments on average and each has an average duration of 8.2 seconds. \emph{QVHighlights~\cite{lei2021detecting}} includes over 10,148 cropped videos (150s long), and each video is annotated with at least one query describing the relevant moments (24.6s in average). In total, there are 10,310 text queries with 18,367 associated moments. Following ~\cite{lei2021detecting,liu2022umt}, we use train split for training and val split for testing. The most commonly used metric for moment retrieval is the average recall at k computed under different temporal Intersection over Union (IoU) thresholds, which is defined as the percentage of at least one of the top-k predicted segments having a larger temporal IoU than the threshold with the ground truth segment, i.e. Recall@K, IoU=[0.5, 0.7]. 


\noindent\textbf{Temporal action localization.} \emph{ActivityNet 1.3}~\cite{heilbron2015activitynet} is a collection of 20,000 untrimmed videos focusing on human actions. Most videos contain only one labeled segment and segments in one video are from the same action class. The dataset is divided into three subsets, train, validation, and test. Following standard practice~\cite{lin2018bsn,lin2019bmn,xu2019gtad,zhang2022actionformer}, we train our models on the training set and report results on the validation set.
The standard evaluation metric for temporal localization is mean Average Precision (mAP) computed under different temporal IoU thresholds. We report mAP under an IoU threshold of 0.5, denoted as mAP@0.5IoU. We also report results for the zero-shot setting, following the data split protocols proposed by~\cite{ju2022prompting,nag2022zero}: 1) training on 50\% of the action labels and testing on the remaining 50\%; 2) training on 75\% of the labels and testing on the rest 25\%. These are created using $10$ random splits of the data, following~\cite{ju2022prompting,nag2022zero}.
In the rest of paper, we use ANet TAL and ANet MR to denote \emph{ActivityNet 1.3} and \emph{ActivityNet Captions}, respectively.

\noindent\textbf{Action segmentation.}
The \emph{COIN}~\cite{tang2019coin} dataset consists of 11,827 training videos and 2,797 testing videos. Each video is labeled with an average of 3.9 segments where each segment lasts 14.9 seconds on average. The segment labels describe a step needed to complete a task, such as ``take out the old bulb'', ``install the new bulb'', etc. Frame accuracy is the primary metric used in the COIN action segmentation task, which is defined as the number of correctly predicted frames divided by the total number of frames. However given how a large proportion of the frames are labelled as background (58.9\%), a naive majority-class prediction model will already get an accuracy of 58.9\% (shown in the first row of Table~\ref{tab:coin}). Hence we also report mean Average Precision (mAP), which averages AP over the classes (excluding background) and is therefore not directly impacted by the large proportion of background. 


\subsubsection{Implementation details}
\label{sec:implementation}
\noindent\textbf{Model Architecture:} 
In \model{}-Base and Large models, the image and text encoders follow the same architecture used in CLIP-B and CLIP-L. The video-text fusion module is implemented using a 6-layer Transformer and the hidden size is set to 512 and 768 for \model{}-B and \model{}-L and the MLP dimension is set to 2048 and 3072, respectively. We construct a 4-layer feature pyramid from the last layer in the video-text fusion module following the procedure described in Section~\ref{sec:fpn}. Following~\cite{zhang2022actionformer}, an output regression range is specified for each level in the pyramid, which is set to [0, 4], [4, 8], [8, 16], [16, $\inf$], respectively ordered from bottom to the top. All heads across different pyramid levels share the same weights, and are randomly initialized. \\
\noindent\textbf{Pretraining:}
Our models are pretrained on Kinetics (K700~\cite{carreira2019short} for our best models, K400 for ablations). The pretraining task is a 400/700-way binary classification problem using a sigmoid cross entropy loss.  For example, for each video we feed all class names into the text tower and the objective is to classify whether or not the video matches any of the class names. During Kinetics pretraining, the image encoder is finetuned and the text encoder is kept frozen to avoid catastrophic forgetting due to the fact that we are finetuning on a small fixed set of vocabulary in Kinetics. The video-text fusion module is always finetuned. \\
\noindent\textbf{Training:}
In training the frames are first resized to have a shorter side of 256 and models are trained on a random crop of size $224\times 224$. For TAL and AS class names are augmented using Kinetics prompts released by~\cite{radford2021learning}, e.g., ``\texttt{a video of a person doing \{label\}}''.  Unless specified otherwise, all TAL and MR models are trained on 128 frames evenly spaced sampled across the whole video. 
This follows the sampling strategy adopted by~\cite{zhang2022actionformer} to deal with videos of varying lengths. Unless specified otherwise, for AS on the COIN dataset, we extract the RGB frames at 2FPS, which is the labelling resolution. We randomly sample 512 consecutive frames and apply padding for videos with less than 512 frames.
All models are trained using synchronous SGD with a momentum of 0.9, with a batch size of $64$.  We follow~\cite{arnab2021vivit} and apply the same data augmentation and regularization schemes~\cite{cubuk_arxiv_2019,huang_stochasticdepth_eccv_2016}, which were used by~\cite{touvron_arxiv_2020} to train vision transformers more effectively. For more implementation details and hyperparameters, we refer readers to the appendix and code.
Our model is implemented using the Scenic library~\cite{dehghani2021scenic} and JAX~\cite{jax2018github}.

\noindent\textbf{Inference} During inference, our results are obtained evaluating a single central crop of $224\times 224$. 
For AS on COIN, we run our model in a non-overlapping sliding window fashion with a window size of 512 frames.
For TAL and AS, we report two results, one using the first prompt and the other by averaging all 28 context prompts, which is defined as prompt ensembling in~\cite{radford2021learning}.

\subsection{Ablations} \label{sec:ablations}
We use the hyperparameters described in Sec.~\ref{sec:implementation} as the default setting for all experiments in the ablation unless specified otherwise. For AS on COIN we randomly sample 128 consecutive frames (instead of 512) for efficiency during training for the ablations. 
For the ablations we report ANet TAL with mAP@0.5IoU, ANet MR with Recall@1 under IoU=0.5 and COIN with mAP.

\input{localization_ablation}
\textbf{Architectural design choices.}
In Table~\ref{tab:localization_ablation}, we ablate three design choices: the loss function, feature pyramid design, and the number of convolution layers in the localization heads. All losses perform similarly with L1 being slightly better than other three. ViTDet-style feature pyramid outperforms a standard FPN~\cite{lin2017feature} as it introduces less additional layers to the pretrained model. Removing the feature pyramid completely significantly degrades the performance, with a 7.4\% drop. Performance increases as we increase the number of convolution layers but saturates at 3. The best setup derived here is used by following experiments.


\textbf{Variations on the text encoder and tokens.}
In Table~\ref{tab:text_encoders}, we freeze the CLIP image encoder, pair it with different text encoders, and finetune them. Using ``unpaired'' image-text encoders indeed diminishes the performance on all three tasks, especially for TAL and MR.
For closed-vocabulary tasks, such as TAL, a text encoder is not strictly required. We hence compare our model to a version without the text encoder, and try to make minimum changes to ensure a fair comparison. 
Without the text tokens the video-text fusion module becomes a temporal encoder ({\ie a transformer which operates on frame-level features, aggregating temporal information across them). To enable this ablation, we also modify the linear projections in Eqs.~\ref{eq:cls_head} and~\ref{eq:reg_head} as follows:
\begin{align*}
\mathbf{\hat{Y}} &= \mathbf{Z}\mathbf{W}_{cls} + \mathbf{b}_{cls} \\
\Delta\mathbf{\hat{T}} &= \text{relu}(\mathbf{Z}\mathbf{W}_{reg} + \mathbf{b}_{reg})
\end{align*}
where $\mathbf{Z} \in \mathbb{R}^{N\times K}$ are the activiations after convolution layers, $\mathbf{\hat{Y}} \in \mathbb{R}^{N \times C}$ and $\Delta\mathbf{\hat{T}} \in \mathbb{R}^{N\times 2C}$ are the predicted class logits and start/end time displacements.

After removing the text encoder, the performance on ANet TAL drops from 54.7 mAP@0.5IoU to 46.5 (a relative decrease of \textbf{15\%}).
In a second study, we also compare the performance of using a single text \texttt{[CLS]} token versus using all the text tokens from the text encoder on different tasks shown in Table~\ref{tab:all_tokens}. For close-vocabulary tasks, such as TAL and AS, \textit{all} refers to 16 tokens to represent the class labels and for MR we increase the sequence length to \textit{32}, i.e., captions contain more words than class labels. We demonstrate that using all tokens gives better performance on all tasks and such improvement is larger for tasks involved more complex language queries, such as MR.

\textbf{Effect of video-text fusion module.}
We also compare our model with a late-fusion variant where the frame relevancy scores are computed as the dot product between the normalized $\mathbf{Z}$ and the class label text embeddings. This variant improves over the no-text variant to 49.8 on ANet TAL but still worse than our proposed mid-fusion model. We find that video-text fusion is essential for achieving good performance on TAL.

\textbf{Finetuning strategies.} Table~\ref{tab:frozen_encoders} compares four different strategies for finetuning a Kinetics-pretrained model on downstream tasks by either freezing or finetuning each of the two encoders. In this study, we always finetune the video-text fusion layers and heads. We observe that it is more beneficial to finetune the image encoder for close-vocabulary tasks, such as TAL and AS. However, for task involving more complex queries, such as MR, finetuning the image encoder actually degrades the performance. A similar phenomenon is also observed by~\cite{zhai2022lit}, and may be due to overfitting.

\input{text_encoders}
\input{all_tokens}
\input{frozen_encoders}



\subsection{Comparison with the state-of-the-art} \label{sec:sota}
In this section we compare to the state-of-the-art for all three tasks individually. Qualitative examples for each task are provided in Fig.~\ref{fig:qualitative}.
\input{mr}
\paragraph{Moment retrieval}
For MR models we freeze the image encoder and finetune the rest of the network following the best strategy derived in Table~\ref{tab:frozen_encoders}. On ANet MR, our \model{}-L model achieves a new state-of-the-art improving the previous best by 2.0\% and 0.4\% in recall@1 under IoU=0.5 and 0.7, respectively (Table~\ref{tab:mr}). On Charades-STA, our \model{}-L model improves upon the previous best~\cite{mun2020local} by 1.3\% and 2.9\% on the same two metrics. On ANet MR, \model{}-L outperforms~\cite{mun2020local} by a larger margin, 6.8\% and 7.1\%. On QVHighlights, UnLoc-
L improves upon the previous best \cite{moon2023query} by 3.7\% and 1.7\%. Most previous work is built upon pre-extracted convolutional features, such as I3D~\cite{carreira_cvpr_2017}, P3D~\cite{qiu2017learning}, C3D~\cite{tran_iccv_2015}, R(2+1)D~\cite{tran_cvpr_2018}, VGG~\cite{simonyan2014very}, SlowFast~\cite{feichtenhofer_iccv_2019}, etc, and our work is most comparable to~\cite{lei2021detecting}, which also employs CLIP features (in addition to SlowFast~\cite{feichtenhofer_iccv_2019} features). Our \model{}-L model scores 5.1\% and 4.4\% higher than~\cite{lei2021detecting} on Charades-STA in recall@1 under IoU=0.5 and 0.7. To the best of our knowledge, we are the first work employing pure transformer features that achieves state-of-the-art results on moment retrieval, which has largely been dominated by CNN-based features.

\input{tal.tex}
\paragraph{Temporal localization}
\input{qualitative}
Table~\ref{tab:tal} shows results on ANet TAL under two settings (finetuned and zero-shot). In the finetuned setting, we freeze the text encoder and finetune the rest of the network following the best strategy derived from Table~\ref{tab:frozen_encoders}. For \model{}-L we increase the sampled frames to 160 and use a 5-L Feature Pyramid. As shown in Table~\ref{tab:tal}, most high-performance methods are built on top of 3D convolutional features. There are two previous attempts to replace the CNN vision encoder by a Transformer encoder. EffPrompt~\cite{ju2022prompting}, built on top of frozen CLIP features, scored significantly lower than recent CNN-based models and STALE~\cite{nag2022zero}, which is also built upon CLIP features, achieved competitive results with the best CNN methods but is 2.2 worse than the same model trained on two-stream I3D features. To the best of our knowledge, we are the first work that achieved state-of-the-art results using only Transformer features. Our \model{}-L model improved previous best results in terms of mAP@0.5IoU by 2.3 and with prompt ensembling this margin is increased to 2.8. 

 For both splits in the zero-shot (open-set) protocols proposed by~\cite{ju2022prompting,nag2022zero}, \model{}-B and L outperform previous best by a significant margin. Specifically, \model{}-L advances previous state-of-the-art by 11.6, a relative 36.1\% improvement on the 50/50 split and by 10.6, a relative 27.7\% on the 75/25 split.

\paragraph{Action segmentation}
\input{coin}

 Table~\ref{tab:coin} compares our model with previous work and \model{}-L outperform previous state-of-the-art by 2.8\% in frame accuracy. Besides architectural differences, we note that previous works~\cite{miech2020end,xu-etal-2021-videoclip,luo2020univl} pretrain their models on HowTo100M~\cite{miech2019howto100m}, which consists of around 100M aligned ASR and video clip pairs, and is also in a similar domain to COIN (instructional web videos). Our models on the other hand, are initialized from CLIP checkpoints, which are trained on cleaner web image-text pairs from multiple domains and finetuned on Kinetics, 10s clips of human activity videos.

%% file: localization_ablation.tex

\begin{table} 
\setlength{\tabcolsep}{4pt}
\centering
\scriptsize{
\begin{tabular}{cccc|ccc|cccc}
    \toprule
    \multicolumn{4}{c|}{Losses}  & \multicolumn{3}{c|}{Feature Pyramid} & \multicolumn{4}{c}{\# conv layers} \\
    \midrule
    L1 & IoU & L1+IoU & DIoU & No & FPN & ViTDet & 1 & 2 & 3 & 4 \\
    \textbf{54.6} & 54.0 & 53.9 & 54.1 & 47.3 & 53.8 & \textbf{54.7} & 52.5 & 53.4 & \textbf{54.7} & 54.5 \\
    \bottomrule
\end{tabular}
}
\vspace{\baselineskip}
\caption{\textbf{Effect of architecture design and losses.} Results are presented on the ANet TAL for mAP@0.5IoU. We compare 4 popular regression losses, two types of feature pyramids (and no pyramid), and the number of convolutional layers in the localization heads. 
} 
\label{tab:localization_ablation}
\end{table}


%% file: text_encoders.tex
\begin{table} \setlength{\tabcolsep}{4pt}
\centering
\scriptsize{
\begin{tabular}{  l  c  c  c c}
    \toprule
    Text Encoder & MParams & ANet TAL & ANet MR & COIN \\
    \midrule
    T5-S & 147.1 & 46.7 & 39.7 & 16.1 \\
    T5-B & 221.5 & 46.6 & 39.9 & 15.9 \\
    CLIP-B & 174.9 & \textbf{53.3} & \textbf{44.2} & \textbf{16.4} \\
    \bottomrule
\end{tabular}
}
\vspace{\baselineskip}
\caption{\textbf{Effect of different text encoders}. We use the same frozen CLIP-B image encoder, with both T5 and CLIP-B text encoders and show results across all tasks. Paired image/text encoders significantly outperform unpaired encoders for localization tasks. Note that for COIN, results are reported using mAP.
}
\label{tab:text_encoders}
\end{table}


%% file: all_tokens.tex
\begin{table} \setlength{\tabcolsep}{4pt}
\centering
\scriptsize{
\begin{tabular}{ l c c c} 
    \toprule
    \# tokens & ANet TAL & ANet MR & COIN \\
    \midrule
    All & \textbf{53.7} & \textbf{44.2} & \textbf{16.4} \\
    One & 53.3 & 42.6 & 15.7 \\
    \bottomrule
\end{tabular}
}
\vspace{\baselineskip}
\caption{\textbf{Effect of number of text tokens.} We show that using all text tokens (16 tokens for both TAL and AS and 32 tokens for MR) performs better than using a single token in video-text fusion on different tasks. Note that the image encoder is frozen.}
\label{tab:all_tokens}
\end{table}


%% file: frozen_encoders.tex
\begin{table} 
\centering
\setlength{\tabcolsep}{4pt}
\scriptsize{
\resizebox{0.8\columnwidth}{!}{%
\begin{tabular}{  l  c  c c}
    \toprule
    Image/Text encoders & ANet TAL & ANet MR & COIN \\
    \midrule
    frozen/frozen & 53.2 & 43.4 & 16.1 \\
    frozen/finetuned & 53.3 & \textbf{44.2} & 16.4 \\
    finetuned/frozen & \textbf{54.7} & 39.7 & 16.6 \\
    finetuned/finetuned & 54.3 & 41.2 & \textbf{16.9} \\
    \bottomrule
\end{tabular}
}
}
\vspace{\baselineskip}
\caption{\textbf{Effect of freezing or finetuning image/text encoder on different tasks.} The video-text fusion module and heads are always finetuned. For closed-vocabulary tasks, such as TAL and AS, finetune the image encoder is better (bottom two rows), however, for tasks involving more complex queries such as MR, finetuning the image encoder degrades performance (top two rows).}
\label{tab:frozen_encoders}
\end{table}


%% file: mr.tex
\begin{table} 
\centering
\setlength{\tabcolsep}{3pt}
\scriptsize{
\begin{tabular}{l l l  c  c  c c}
    \toprule
    & \multirow{2}{*}{Method} & \multirow{2}{*}{Vision Enc.} & \multicolumn{2}{c}{R@1} & \multicolumn{2}{c}{R@5} \\
    & & & IoU=0.5 & IoU=0.7 & IoU=0.5 & IoU=0.7 \\
    \midrule
    \parbox[t]{2mm}{\multirow{9}{*}{\rotatebox[origin=c]{90}{Charades-STA}}}
    & CTRL~\cite{gao2017tall} & C3D & 23.6 & 8.9 & 58.9 & 29.5 \\
    & 2D TAN~\cite{zhang2020learning} & VGG & 39.7 & 23.3 & 80.3 & 51.3 \\
    & VSLNet~\cite{zhang2020span} & I3D & 47.3 & 30.2 & - & - \\
    & UMT~\cite{liu2022umt} & VGG & 49.4 & 26.2 & \textbf{89.4} & 55.0 \\
    & IVG-DCL~\cite{nan2021interventional} & C3D & 50.2 & 32.9 & - & - \\
    & M-DETR~\cite{lei2021detecting} & CLIP & 55.7 & 34.2 & - & - \\
    & LGI~\cite{mun2020local} & I3D & 59.5 & 35.5 & - & - \\ \cmidrule{2-7}
    & \model{}-B & CLIP & 58.1 & 35.4 & 87.4 & \textbf{59.1} \\
    & \model{}-L & CLIP & \textbf{60.8} & \textbf{38.4} & 88.2 & \textbf{61.1} \\
    \midrule 
    \parbox[t]{2mm}{\multirow{7}{*}{\rotatebox[origin=c]{90}{ANet MR}}} 
    & LGI~\cite{mun2020local} & C3D & 41.5 & 23.1 & - & - \\
    & VSLNet~\cite{zhang2020span} & I3D & 43.2 & 26.2 & - & - \\
    & 2D TAN~\cite{zhang2020learning} & C3D & 44.5 & 26.5 & 77.1 & 62.0 \\
    & DRN~\cite{zeng2020dense} & C3D & 45.5 & 24.4 & 78.0 & 50.3 \\
    & VLG~\cite{soldan2021vlg} & C3D & 46.3 & 29.8 & 77.2 & \textbf{63.3} \\ \cmidrule{2-7}
    & \model{}-B  & CLIP & \textbf{48.0} & 29.7 & \textbf{81.5} & 61.4 \\
    & \model{}-L & CLIP & \textbf{48.3} & \textbf{30.2} & \textbf{79.2} & 61.3 \\  
    \midrule
    \parbox[t]{2mm}{\multirow{5}{*}{\rotatebox[origin=c]{90}{QVHighlights}}} 
    & M-DETR~\cite{lei2021detecting} & SF+CLIP & 53.9 & 34.8 & - & - \\
    & UMT~\cite{liu2022umt} & SF+CLIP & 60.3 & 44.3 & - & - \\
    & QD-DETR~\cite{moon2023query} & SF+CLIP & 62.4 & 45.0 & - & - \\
    \cmidrule{2-7}
    & \model{}-B  & CLIP & \textbf{64.5} & \textbf{48.8} & - & - \\
    & \model{}-L & CLIP & \textbf{66.1} & \textbf{46.7} & - & - \\ 
    \bottomrule
\end{tabular}
}
\vspace{0.4\baselineskip}
\caption{\textbf{Comparison with the state-of-the-art for Moment Retrieval.} We show results on Charades-STA (test split), ANet MR (val\_2 split), and QVHighlights (val split) datasets. 
}
\label{tab:mr}
\end{table}


%% file: tal.tex
\begin{table} 
\centering
\setlength{\tabcolsep}{4pt}
\scriptsize{
\begin{tabular}{l l l c}
    \toprule
    Setting & Method & Vision Encoder & mAP@0.5IoU \\
    \midrule
    \multirow{15}{*}{Finetuned} & A2Net~\cite{yang2020revisiting} & I3D & 43.6 \\
    & TSP~\cite{alwassel2021tsp} & R(2+1)D & 51.3 \\
    & GTAN~\cite{long2019gaussian} & P3D & 52.6 \\
    & VSGN~\cite{zhao2021video} & I3D & 53.3 \\
    & TadTR~\cite{liu2022end} & R(2+1)D & 53.6 \\
    & PBRNet~\cite{liu2020progressive} & I3D & 54.0 \\
    & TCANet~\cite{qing2021temporal} & SlowFast & 54.3 \\
    & ActionFormer~\cite{zhang2022actionformer} & R(2+1)D & 54.7 \\
    & ContextLoc~\cite{zhu2021enriching} & I3D & 56.0 \\
    & EffPrompt~\cite{ju2022prompting} & CLIP & 44.0 \\
    & STALE~\cite{nag2022zero} & CLIP & 54.3 \\
    & STALE~\cite{nag2022zero} & I3D & 56.5 \\ \cmidrule{2-4}
    & \model{}-B (1st prompt) & CLIP & 54.6 \\
    & \model{}-L (1st prompt) & CLIP & \textbf{58.8} \\
    & \model{}-L (prompt ensembling) & CLIP & \textbf{59.3} \\
    \midrule
    \multirow{5}{1.4cm}{Zero-shot 50\% Seen 50\% Unseen} & EffPrompt~\cite{ju2022prompting} & CLIP & 32.0 \\
    & STALE~\cite{nag2022zero} & CLIP & 32.1 \\ \cmidrule{2-4}
    & \model{}-B (1st prompt) & CLIP & \textbf{36.9} \\ 
    & \model{}-L (1st prompt) & CLIP & \textbf{43.2} \\ 
    & \model{}-L (prompt ensembling) & CLIP & \textbf{43.7} \\ 
    \midrule
    \multirow{5}{1.4cm}{Zero-shot 75\% Seen 25\% Unseen} & EffPrompt~\cite{ju2022prompting} & CLIP & 37.6 \\
    & STALE~\cite{nag2022zero} & CLIP & 38.2 \\ \cmidrule{2-4}
    & \model{}-B (1st prompt) & CLIP & \textbf{40.2} \\
    & \model{}-L (1st prompt) & CLIP & \textbf{47.4} \\ 
    & \model{}-L (prompt ensembling) & CLIP & \textbf{48.8} \\
    \bottomrule
\end{tabular}
}
\caption{\textbf{Comparison with the state-of-the-art on ANet TAL.} We show results for finetuning, and both the zero-shot (open-set) protocols introduced by~\cite{ju2022prompting}. Our method outperforms all previous work across all settings, achieving strong gains particularly in the zero-shot settings.}
\label{tab:tal}
\end{table}


%% file: qualitative.tex
\begin{figure*}[t]
    \centering
    \includegraphics[width=0.98\linewidth]{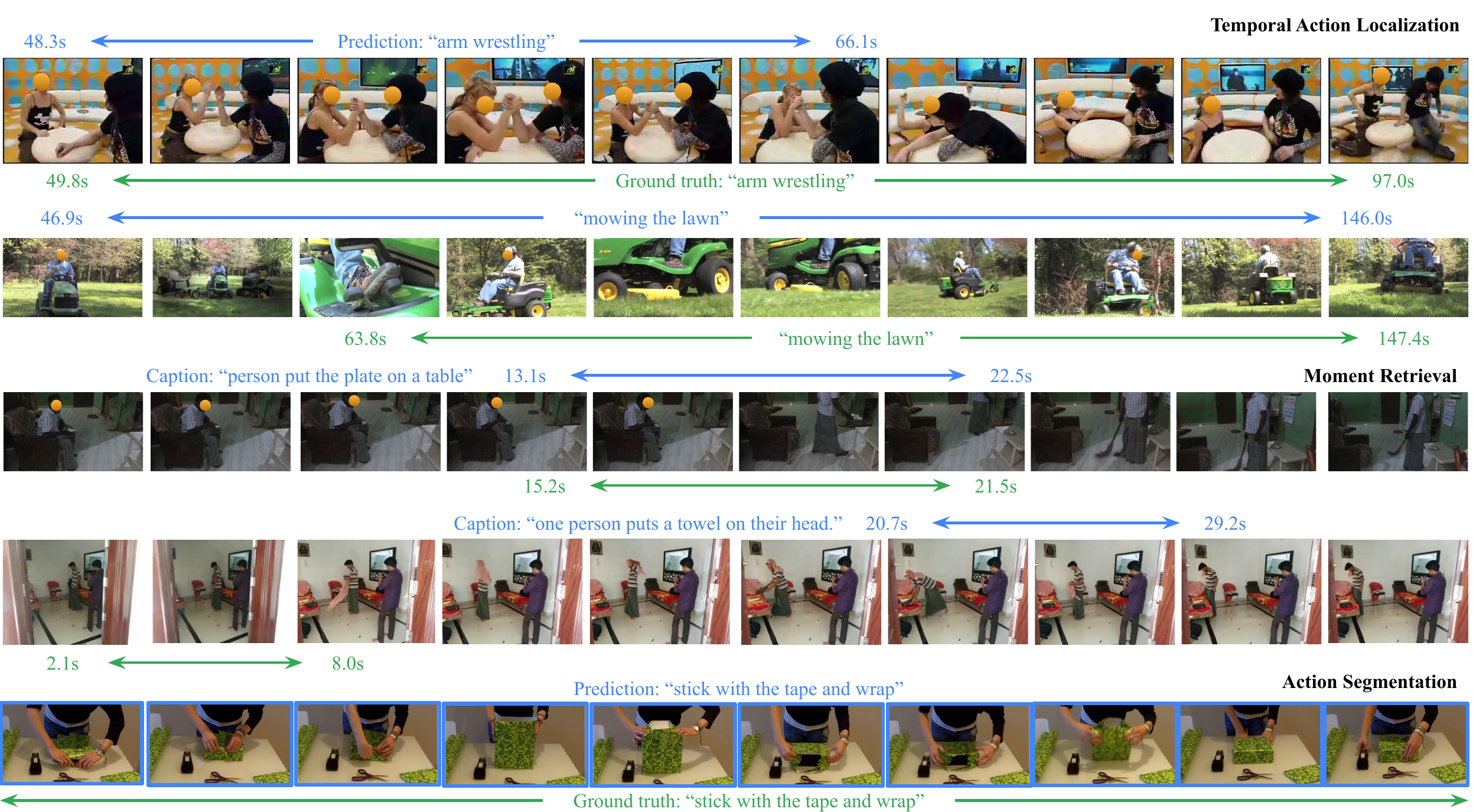}
    \caption{\textbf{Qualitative Results} We show results on ActivityNet, Charades and COIN, for Temporal Action Localization, Moment Retrieval and Action Segmentation respectively. Predictions are shown in blue, while the ground truth is in green (best viewed in colour). For action segmentation, the ground truth covers the entire clip. Note how our model is able to predict accurate boundaries, in some cases better refined than the ground truth (top row, the arm wrestling action has stopped, however the ground truth boundary extends for a while after). For the second example for Moment Retrieval (4th row from top), we show a failure case, where our model detects the moment where the towel is `put down', and not `on their head' as perhaps the latter is a rarer occurrence in the training data.}
    \label{fig:qualitative}
\end{figure*}

%% file: coin.tex
\begin{table} \setlength{\tabcolsep}{4pt}
\centering
\scriptsize{
\begin{tabular}{  l  c  c  c c}
    \toprule
    Method & Frame accuracy & mAP \\
    \midrule
    Baseline: predict all background & 58.9 & 0.0 \\ \cmidrule{1-3}
    ActBERT~\cite{zhu2020actbert} & 57.0 & - \\
    MIL-NCE~\cite{miech2020end} & 61.0 & - \\
    TACo~\cite{yang2021taco} & 68.4 & - \\
    VLM~\cite{xu-etal-2021-vlm} & 68.4 & - \\
    VideoCLIP~\cite{xu-etal-2021-videoclip} & 68.7 & - \\
    UniVL~\cite{luo2020univl} & 70.0 & - \\
    \midrule
    \model{}-B (1st prompt) & 68.0 & 36.2 \\
    \model{}-L (1st prompt) & \textbf{72.6} & 47.0 \\
    \model{}-L (prompt ensembling) & \textbf{72.8} & 47.7 \\
    \bottomrule
\end{tabular}
}
\vspace{\baselineskip}
\caption{\textbf{Comparison with the state-of-the-art on COIN for Action Segmentation.} We report results using both frame accuracy (as is standard practice) and mAP, which we believe is a better metric given that a large proportion (58.9\%) of the dataset is labelled as a single class (background).}
\label{tab:coin}
\end{table}

%% file: conclusion.tex
\section{Conclusion and Future Work} We propose a new model for video localization tasks, called \model{}. \model{} consists of a two-tower CLIP model, the output features of which are fed into a video-text fusion module and feature pyramid. Unlike previous works, we achieve state-of-the-art results on 3 different benchmarks (moment retrieval, temporal action localization and action segmentation) with a single approach, without the need for action proposals or pretrained video features.

Future work will investigate cotraining on the three localization tasks, pretraining on large, weakly labelled datasets, 
exploring highlight detection as an additional downstream task, 
and adapting our model to other modalities such as audio for sound localization~\cite{huh2023epic}. 

%% file: ack.tex
\paragraph{Acknowledgements.}
We thank Xingyi Zhou for helpful discussion and feedback.